\pdfoutput=1

\documentclass[11pt]{article}

\usepackage{acl}

\usepackage{tabularx}  
\usepackage{times}
\usepackage{latexsym}
\usepackage{graphicx}
\usepackage{amsmath}
\usepackage{amssymb}
\usepackage[linesnumbered,ruled,vlined]{algorithm2e}
\usepackage{amsmath, amssymb} 
\usepackage{booktabs, multirow} 
\usepackage{soul}
\usepackage{changepage,threeparttable} 

\usepackage[T1]{fontenc}
\usepackage{xcolor}
\usepackage{rotating}

\usepackage[utf8]{inputenc}

\usepackage{microtype}

\title{Introducing Syllable Tokenization for Low-resource Languages: \\
A Case Study with Swahili}

\author{Jesse Atuhurra, Hiroyuki Shindo, Hidetaka Kamigaito, Taro Watanabe\\ \AND
  \texttt{Division of Information Science, NAIST} \\ 
  \footnotesize
  {
  \texttt{ \{atuhurra.jesse.ag2, shindo, kamigaito.h, taro\} @naist.ac.jp} 
  } \\ }

\begin{document}
\maketitle
\begin{abstract}
In multilingual NLP, many attempts have been made to ensure that pre-trained language models (e.g., mBERT, GPT-2) get better and become applicable to low-resource languages. To achieve multilingualism for pre-trained language models (PLMs), there’s a need for techniques to create word embeddings that capture the linguistic characteristics of any language. Tokenization is one such technique because it allows for the words to be split based on characters or sub-words, creating word embeddings that best represent the structure of the language. Creating such word embeddings is essential to applying PLMs to other languages where the model was not trained, enabling multilingual NLP. However, most PLMs use generic tokenization methods (e.g., BPE, wordpiece, unigram), which may not suit specific languages. We hypothesize that tokenization based on syllables within the input text (which we call \texttt{syllable tokenization}) should facilitate the development of syllable-aware language models. The syllable-aware language models make it possible to apply PLMs to languages that are rich in syllables, e.g., Swahili. Previous works introduced subword tokenization. Our work extends such efforts. Notably, we propose a \texttt{syllable tokenizer} and adopt an experiment-centric approach to validate the proposed tokenizer based on the Swahili language. We conducted text generation experiments with GPT-2 to evaluate the effectiveness of the syllable tokenizer. Our results show that the proposed syllable tokenizer generates syllable embeddings that effectively represent the Swahili language. 
\end{abstract}
\begin{figure}[t!]
\centering
\includegraphics[width=7.5cm]{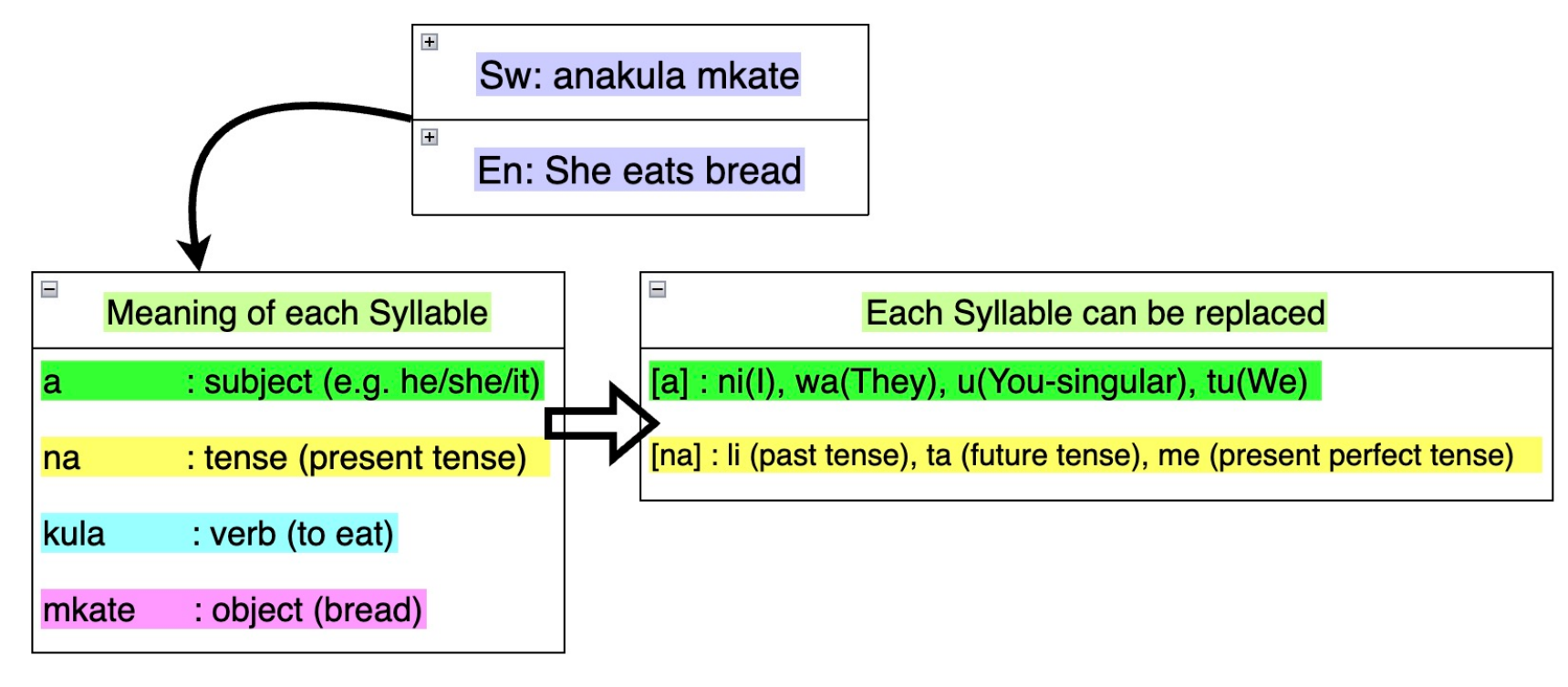}
\caption{Example of a Swahili sentence (Sw) and the translation in English (En). The word "anakula" is broken down into syllables and explained in the bottom left table. We can replace the syllables, with syllables on the bottom right, creating new meanings. For example, the subject ``a'' can be replaced with ``wa'' resulting in ``wanakula'' to mean ``They eat bread.''. Similarly, the tense ``na'' can be replaced with ``li'' resulting in ``alikula'' to mean ``She ate bread.'' This example demonstrates the richness of syllables in Swahili.}
\label{fig:tokenized_sentence}
\end{figure}
\begin{table*}[!t]
\tiny
\centering
\begin{tabularx}{0.8\linewidth}{*{13}{>{\centering\arraybackslash}X}} 
    \toprule
    \multicolumn{13}{c}{ {\bf Swahili Syllabic Alphabet}   }\\
    \hline
    \hline
        mbwa  & mbwe & mbwi & ndwa & ndwe &ndwi & ngwa & ngwe&ngwi& njwa& njwe& njwi& nywa\\
        nywe & shwa & shwe & shwi & chwa&chwe &chwi &pwa &pwe &pwi &pwo &swa &swe \\
        swi & twa & twe & twi & zwa& zwe &zwi &cha&che &chi &cho &chu &dha\\
        dhe & dhi & dho & dhu & gha&ghe &ghi &gho &ghu &kha &khe &kho &khu\\
        mba & mbe & mbi & mbo & mbu& nda& nde&ndi &ndo &ndu &nga &nge& ngi\\
        ngo & ngu & ng’a & ng’e & ng’o& nja& nje& nji &njo&nju& nya &nye &nyi\\
        nyo & nyu & sha & she & shi&sho &shu &tha &the& thi& tho& thu& vya\\
        vye & vyo & bwa & bwe & bwi&gwa &gwe& gwi &jwa &jwe& jwi& kwa& kwe\\
        kwi & lwa & lwe & lwi & mwa & mwe & mwi & nza & nze & nzi & nzo & nzu &ba\\
        be &  bi & bo & bu & da & de & di & do & du & fa & fe & fi & fo\\
        fu & ga & ge & gi & go & gu & ha & he & hi & ho & hu & ja & je\\
        ji & jo & ju & ka & ke & ki & ko & ku & la & le & li & lo & lu\\
        ma & me & mi & mo & mu & na & ne & ni & no & nu & pa & pe & pi\\
        po & pu & ra & re & ri&ro &ru& sa& se& si& so& su& ta\\
        te & ti & to & va & ve & vi & vo & vu & wa & we & wi & wo\\
        wu & ya & ye & yi & yo & yu & vu & za & ze & zi & zo & zu & a\\
        e & i & o & u & b & d & f & k & m & n & s & - & -\\
    \bottomrule
\end{tabularx}
\caption{List of Swahili syllables. There are 219 syllables~\cite{app9183648} in total.}
\label{table:ListOfSwahiliSyllables}
\end{table*}
\section{Introduction} \label{introduction}
Multilingual NLP is essential for the success of NLP, in general. The ability of contemporary language models, PLMs, to generalize to resource-scarce languages is beneficial to the NLP research community. It could have much greater merits, like preserving extinct or endangered languages. Resource-scarce languages in an NLP context tend to reflect the scarcity of the resources in the community from which those languages originate. For machine learning researchers, “low-resourced” identifies languages for which few digital or computational data resources exist, often classified in comparison to another language~\cite{gu-etal-2018-universal},\cite{zoph-etal-2016-transfer}. On the contrary, to the sociolinguist, 
“low-resourced” can be broken down into many categories: low density, less commonly taught, or endangered, each carrying slightly different meanings~\cite{cieri-etal-2016-selection}.

To achieve multilingualism for pre-trained language models (PLMs), there’s a need for techniques to create word embeddings that capture the linguistic characteristics of any language. One of the essential techniques is tokenization. The tokenizer used in a PLM splits words based on characters, or sub-words, to create word embeddings. Creating word embeddings is critical to training the PLM to perform language tasks such as text generation in any language. Therefore, it is vital to choose the tokenization that best represents the linguistic properties of the target language. Multilingual NLP is successful when the PLM performs tasks in a language that was not trained. 

Moreover, there are other research efforts to develop technologies for low-resource languages, such as those spoken in Africa. Among the notable efforts are:~\cite{siminyu20201st}, ~\cite{emnlp-2019-deep}, and others. In addition, the emergence of multilingual language models (MLMs) and multilingual datasets can not be understated. 

Despite the progress made in Multilingual NLP, most PLMs use generic tokenization methods (e.g., wordpiece, unigram, and byte-pair encoding or BPE), which may only be suitable for some languages. For example, we hypothesize that tokenization based on syllables in the input text can facilitate the development of syllable-aware language models. The syllable-aware language models make it possible to apply PLMs to languages that are rich in syllables, e.g., Swahili. We drew inspiration from recent publications such as;  \cite{nekoto2020participatory}, ~\cite{adelani2021masakhaner}, who extensively studied machine translation and named entity recognition for African languages, and ~\cite{app9183648} who derived word embeddings for Swahili from syllables by use of a convolutional neural network (CNN). Our proposed tokenizer requires no particular neural network, and it splits words into their syllable constituents. In what follows, we call such tokenization \texttt{syllable tokenization} and the corresponding tokenizer, the \texttt{syllable tokenizer}. This tokenizer is meant for syllabic-based and agglutinative languages—to improve language representation for the mainstream PLMs in use today. 

To validate the tokenizer proposed above, we chose Swahili for our study and then conducted experiments to generate Swahili text with GPT-2. We chose Swahili because it is the most spoken language in Africa \cite{570887}, has about 98 million speakers, and is rich in syllables. 

The alphabet for Swahili is based on the Latin alphabet and consists of five vowels (a, e, i, o, u) and twenty-five consonants (b, c, d, f, g, h, j, k, l, m, n, p, r, s, t, v, w, y, z, ch, dh, gh, ng’, sh, th). Hence, one syllable consists of either one vowel or a combination of one consonant and a vowel. For example, the verb \texttt{kula} (to eat), shown in Figure \ref{fig:tokenized_sentence}, can be broken down into two syllables: \texttt{ku} and \texttt{la}. Swahili is highly agglutinative, and it has many polysemous features. Moreover, Swahili morphology depends on prefixes and suffixes, which are syllables. As a result, the position of each syllable in the word also bears syntactic and semantic meaning. Such syllable information cannot be represented accurately by current tokenization methods (such as BPE and wordpiece which are the default tokenization in GPT-2 and mBERT respectively) used in PLMs, necessitating a different tokenization approach. 

To summarize, we developed a new tokenizer called the  \texttt{syllable tokenizer} and measured its effectiveness by investigating the quality of Swahili text generated by GPT-2. Our main contributions include: 
\begin{itemize}
    \item We developed a syllable tokenizer to effectively split words in Swahili text into syllable tokens and create syllable embeddings. 
    \item We used above syllable embeddings as input to GPT-2 to generate new text with GPT-2. 
\end{itemize}

We leave the analysis of the syllable tokenizer on other low-resource languages (especially African languages) for future work due to differences in morphologies, a non-overlapping set of rules, syllables, vocabulary, and words, among others. In particular, when compared to Gikuyu, Kinyarwanda, Lingala, Haya, and Luganda (which are spoken in Kenya, Rwanda, Democratic Republic of Congo, Tanzania, and Uganda, respectively), Swahili essentially differs in phonological inventory because Swahili consists of a more extensive inventory of consonants, a smaller inventory of vowels, and irregularities related to stress and syllabification \cite{Jerro2018LinguisticCA}.
\section{Related Work} 
\label{Related Work}
Our work is motivated by ~\cite{choi-etal-2017-syllable}, ~\cite{yu-etal-2017-syllable},~\cite{Ander_Martinez2021} and ~\cite{app9183648} who proposed word embeddings from syllable embeddings (WEFSE) for Swahili. They showed that such embeddings were better suited for the creation of a syllable-aware language model than word2vec. Unlike ~\cite{app9183648} who created syllable embeddings from convolutional neural networks, we deviate from this approach and generate syllable embeddings following the linguistic structure of Swahili, wherein a syllable comprises one vowel or a a combination of one vowel and one consonant. Recent progress on natural language generation (NLG) especially in English makes it very tempting to make similar progress in other languages, such as Swahili. We performed text-generation using GPT-2, and tokenization plays such an important role in capturing the linguistic properties of the language. GPT-2~\cite{radford2019language} is a transformer-based language model with 1.5 billion parameters trained on a dataset of 8 million web pages. GPT-2 is trained with an objective of predicting the next word, given all of the previous words within a text. GPT-2 employs byte-level Byte-Pair Encoding (BPE) tokenization in which the base vocabulary (of size 256) consists of bytes, and each base character is included in the vocabulary. Eventually, GPT-2 has a vocabulary size of 50,257\footnote{A summary on tokenizers can be found here https://huggingface.co/docs/transformers/tokenizer\_summary}. In general, tokenization involves splitting text into words or sub-words. Major tokenization techniques include tokenization based on: space, punctuation, rules, characters and unigrams. However, most transformers use sub-word tokenization which incorporates both word-level and character-level tokenization with the main aim being that frequently used words should not be split into smaller sub-words, but rare words should be decomposed into meaningful sub-words. Sub-word tokenization examples are: Byte-level BPE (used in GPT-2), and WordPiece (used in BERT). WordPiece tokenization, similar to BPE, progressively merges characters from the base vocabulary but differs from BPE in the sense that WordPiece tokenization selects the pair of symbols that will maximize the likelihood of training data if added to the vocabulary. It is clear that both BPE and WordPiece do not split words based on syllables hence syllabic information is lost. We hypothesize that such tokenization is insufficient for developing syllable-aware language models.  Other notable tokenizers include: jieba\footnote{https://github.com/fxsjy/jieba} for Chinese, MeCab\footnote{https://github.com/jordwest/mecab-docs-en} for Japanese and Morfessor\footnote{https://morfessor.readthedocs.io/en/latest/}. We compared two algorithms, BPE, WordPiece which were used in GPT-2/BERT as reference implementations; with the proposed \texttt{syllable tokenizer}.

\section{Method and Motivation} \label{Method and Motivation}

\textbf{\textit{Motivation:}} In general, we hypothesize that splitting words based on language features (e.g., syllables, as shown in Figure \ref{fig:Syllable}) is better than splitting words based on statistical features. We aim to compare the Syllable tokenizer which is language-feature based, to the BERT tokenizer and the BPE tokenizer, which are statistical-based.

We investigated three different tokenization strategies. Syllable tokenization includes syllable features e.g., one syllable has one meaning. BPE tokenization has the same granularity as the original GPT-2 pretrained  model but has been fine-tuned on a different language (i.e., Swahili language but not the English language). The tokenization in BERT employs the WordPiece sub-word tokenization algorithm.

\textbf{\textit{Syllable Tokenizer:}} In this paper, we propose a novel tokenizer based on syllables contained in text. The Syllable tokenizer is motivated by the finite set of syllables available in Swahili~\cite{app9183648}, or other syllabic-rich languages. The syllables are shown in Table \ref{table:ListOfSwahiliSyllables}. There are two hundred nineteen (219) syllables in total.

Let \( \mathbb{Y} \) be the set of all syllables available in Swahili such that \( \mathbb{Y} = \{y_1, y_2, \ldots, y_n \} \) where \( y_i \) is a syllable as shown in Table \ref{table:ListOfSwahiliSyllables}. 

Any syllable \(y_i\) is an element of set \(\mathbb{Y}\), where the set \(\mathbb{Y} \in \mathbb{R}^{219}\).

During syllable tokenization, Swahili words contained in the text corpus are split first on whitespace, and then on vowels, as shown in the Algorithm \ref{algorithm:Tokenization}, resulting in several syllable tokens $y_i$. Therefore the vocabulary size is 219, that is, the set of all elements in $\mathbb{Y}$. This significantly improves the computation efficiency because the vocabulary size is very small.

Given a Swahili sentence \(\mathbb{X}_i\) containing \(m\) number of syllables, the effective matrix dimension is \(\mathbb{X}_i \in \mathbb{R}^{219 \times m}\).
\begin{algorithm}[t]
\DontPrintSemicolon 
\SetKwData{Left}{left}\SetKwData{This}{this}\SetKwData{Up}{up}
\SetKwFunction{Union}{Union}\SetKwFunction{FindCompress}{FindCompress}
\SetKwInOut{Input}{input}\SetKwInOut{Output}{output}

\KwData{Set of all syllables: $\mathbb{Y} = \{y_1, y_2, \ldots, y_n \}$ and $\mathbb{Y} \in \mathbb{R}^{219}$}
\KwData{Set of words in sentence: $\mathbb{X} = \{x_1, x_2, \ldots, x_m \}$}
\KwResult{A set of syllables in $\mathbb{X}$ from $\mathbb{Y}$}
\textbf{Initialization:} $t=0$\;
$l = 0$\; 
\While{$l \neq$ \text{length}($\mathbb{X}$)}{
  Split all words on whitespace\;
  Remove all unknown characters\;
  \eIf{$l \leq$ \text{len}($\mathbb{X}$)}{
   Iterate over words $x_i$ in sentence $\mathbb{X}$\;
   Split words after each vowel\;
   Starting from left to right\;
   Return sequence of syllable tokens $x_1, x_2, x_3, \ldots, x_m$\;
   End of the sentence\;
   Move to next sentence\;
   Repeat all processes above\;
   }{
   No sentences available\;
   Stop\;
  }

}
\label{algorithm:Tokenization}
\caption{Syllable Tokenization for Swahili.}
\end{algorithm}
\section{Experimental Settings} \label{Experimental Settings}
\subsection{Datasets}
In this work, all the evaluation is based mainly on the Swahili language. 
We used Swahili Wikipedia articles available at Wikimedia\footnote{https://dumps.wikimedia.org/swwiki}, for training the GPT-2 during the text-generation task. In order to do more elaborate analysis, we also used the Helsinki Corpus of Swahili 2.0 (HCS 2.0), which is provided by the Language bank of Finland\footnote{https://metashare.csc.fi/repository/browse/helsinki-corpus-of-swahili-20-hcs-20-annotated-version}.
\begin{table}[t]
  \centering
  \begin{tabular}{lr}
  \toprule
     Split & \# Sentences\\
  \hline
  \hline
    Training set  & 272,934\\
    Testing set & 30,326 \\
    \hline
    Total & 303,260\\
  \bottomrule
  \end{tabular}
  \caption{ The number of sentences in the Training and Testing sets.}
  \label{table:Number of sentences.}
 \end{table}
As shown in Table \ref{table:Number of sentences.}, we used a 90/10 split for all the data. Our Swahili data contains 303k sentences in total.
\begin{figure*}[t]
  \centering
  \includegraphics[width=\textwidth]{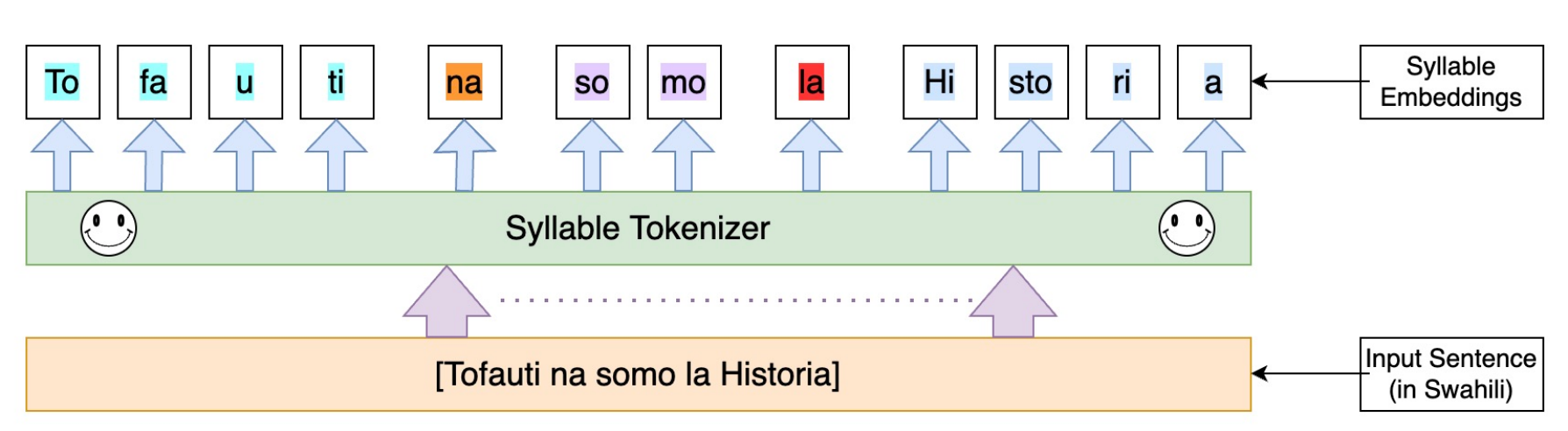}
  \caption{Segmentation of Swahili words with a Syllable Tokenizer. There are five words in the input sentence. After segmentation, syllables indicated with the same color originate from the same word.}%
  \label{fig:Syllable}
\end{figure*}
\subsection{Implementation Details}
We adopt the standard GPT-2~\cite{radford2019language} implementation available at HuggingFace \footnote{https://huggingface.co/transformers/v3.5.1/model\_doc/gpt2.html}. We set the tokenizer in GPT-2 to be our novel Syllable tokenizer (instead of the \textit{GPT2Tokenizer}) and fine-tuned GPT-2 on a training set consisting of Swahili sentences, to generate new Swahili sentences.  We used the Adam optimizer with a learning rate of $5\mathrm{e}{-4}$  and a linear decay learning rate scheduler with $1\mathrm{e}{+2}$ linear warmup steps during fine-tuning. Due to memory constraints, the batch size was set to 2 and the training was done for 5 epochs. Evaluation was done for every 100 steps on the Swahili validation set to select the best model. To ensure reproducibility, we used a a random seed of 42. The dimensionality of the embeddings and hidden states was 768. The vocabulary size of the standard GPT-2 model is 50,257 and the number of hidden layers in the transformer encoder is 12. We show examples of Swahili sentences generated by GPT-2 in the Appendix \ref{sec:appendix}.
\subsection{Text generation with GPT-2}
We fine-tuned GPT-2 to generate Swahili sentences. The assumption is that GPT-2 has large enough parameters within the language model head, necessary to generalize well to any language; in a zero-shot setting. Using GPT-2 as our language model, we train this model on Swahili corpora. We investigate; with different tokenization algorithms i.e., Byte-level BPE, WordPiece, and Syllable-based;—the impact of tokenization schemes on the quality of Swahili text generated. The generated Swahili text is evaluated by a native Swahili speaker. The human evaluation analyses the: Language structure, Meaning, Word order and Fluency; each on a Likert scale of 1 to 5. The purpose of this experiment is to: \\
(a) Analyse how well GPT-2 performs on low-resource languages such as Swahili.\\
(b) Investigate the effect of tokenization on the quality of text generated by GPT-2.
\section{Results and Discussions} \label{Results and Discussions}
\subsection{Syllable tokenizer}
The main aim of the Syllable tokenizer is to capture the syllable representation in Swahili language. As shown in Figure~\ref{fig:Syllable}, the proposed Syllable tokenizer correctly split the sentence \texttt{Tofauti na somo la Historia} into constituent syllables following the rules described in Algorithm \ref{algorithm:Tokenization}.
\subsection{Swahili text generation}
 \begin{table}[t]
  \centering
  \resizebox{\linewidth}{!}{
  \begin{tabular}{lllll}
  \toprule
    Tokenizer & Structure & Meaning & Word Order & Fluency\\
  \hline
  \hline
    BPE  & 3.27 & 3.43 & 3.28 & 3.46\\
    WordPiece & 4.49 & 4.53 & 4.52 & 4.69 \\
    Syllable & 4.60 & 4.59 & 4.63 & 4.76\\
  \bottomrule
  \end{tabular}}
  \caption {Average scores on each of the four language properties evaluated. We can see that the WordPiece tokenizer has a point higher than the default BPE tokenizer found in GPT2.}
  \label{table:Tokenizer_Scores}
\end{table}
We began by computing the perplexity score on the Swahili Wikipedia articles, under the GPT-2 language model with GPT2TokenizerFast tokenizer. The perplexity score is 19.5885, noting that the default tokenizer in GPT-2 is byte-level BPE. Further experiments were carried out to investigate the impact of tokenization on the quality of Swahili text. Each of BPE, WordPiece, and Syllable tokenization was investigated. The Swahili sentences generated by employing the default byte-level BPE tokenization available in GPT-2 were compared to the WordPiece tokenization used in the BERT tokenizer and the syllable tokenizer.\\

Human evaluation was done by a native Swahili speaker. In Table~\ref{table:Tokenizer_Scores}, it is clear that the new sentences generated by GPT-2 when a Syllable tokenizer is used are more fluent, and those sentences tend to follow the structure and word order of Swahili more better, than when a BPE (\textit{GPT2Tokenizer}) or a WordPiece(\textit{BertTokenizer}) is used.  More findings from the evaluation are described next, and relevant examples are described in the Appendix \ref{sec:appendix}.
\begin{itemize}
    \item Despite the fact that byte-level BPE is the default tokenizer in GPT-2, it was observed that a change of the tokenizer from the default BPE to the Syllable tokenizer improved the quality of the generated Swahili text.
    \item All tokenizers perform well on short sentences (i.e., those with less than 12 words) though BERT tokenizer yields more natural text.
    \item On longer sentences, there remains a challenge with "connecting" ideas. It was observed that often times, when two or more sentences are generated, the sentences tend to be unrelated which gives rise to a clueless or meaningless story. 
    \item BPE tokenizer sometimes struggled with  usage of the correct particles but the BERT tokenizer was able to get particles right in most of the generated Swahili texts. 
    \item In terms of chronological ordering of sentences/ideas, BERT tokenizer produced better output than the BPE tokenizer. 
    \item In a few situations, both the BPE and the BERT tokenizers struggled to use the correct singular/plural forms of nouns. 
    \item Both the BPE and the BERT tokenizers  generated multi-domain stories spanning several spheres of life such as; politics, sports, technology,history, religion, finance, etc.
\end{itemize}
\section{Conclusion} \label{Conclusion}
We proposed a syllable tokenizer and investigated the ability of syllable tokenization (and BPE, wordpiece tokenization) to learn effective word embeddings for Swahili, by conducting text generation experiments with GPT-2. Based on these experiments, syllable tokenization yields better results than other tokenizers. The results reinforce our hypothesis that word embeddings based on syllables, especially for a syllabic-based language, can complement existing approaches on word embeddings, further improving the performance of mainstream NLP models on low-resource languages, such as Swahili. 

This work is only a first step in analyzing syllable tokenization for other syllabic-based languages, such as Niger-Congo or Indic languages. Due to differences in morphologies and a non-overlapping set of rules, syllables, vocabulary, and words, among other factors, we leave the analysis of the syllable tokenizer on other low-resource languages (especially African languages) for future work.
\bibliography{main}

\appendix
\label{sec:appendix}
\section{Appendix}
\subsection{Word segmentation}
\begin{figure*}[!t]%
  \centering
  \includegraphics[width=\textwidth]{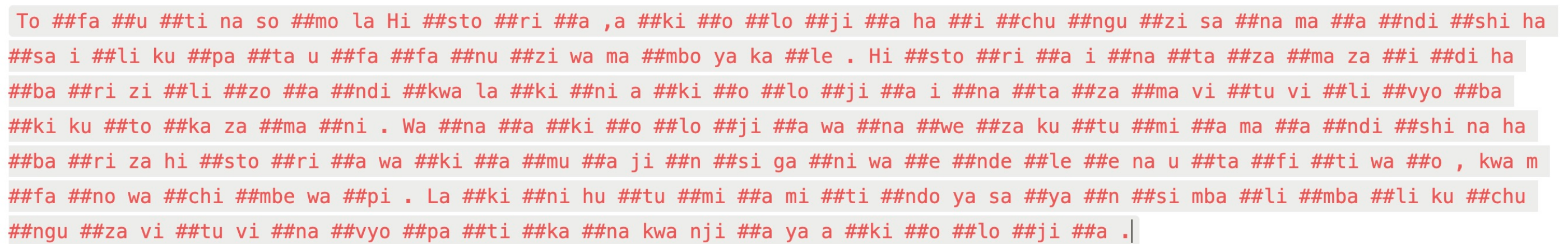}
  \caption{Segmentation of Swahili words with a Syllable Tokenizer.}%
  \label{fig:syllable_segments}
\end{figure*}

\begin{figure*}[!t]%
  \centering
  \includegraphics[width=\textwidth]{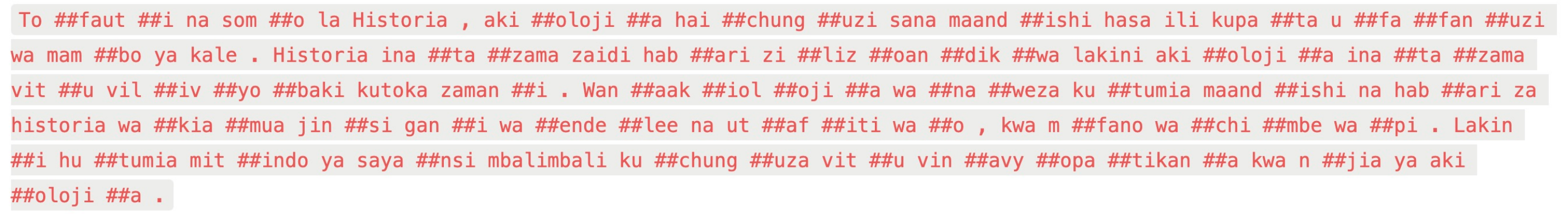}
  \caption{Segmentation of Swahili words with a BERT Tokenizer.}%
  \label{fig:bert_segments}
\end{figure*}

We have shown an example of segmentation based on both the Syllable and the BERT tokenizers. Given the text: \texttt{``Tofauti na somo la Historia,akiolojia haichunguzi sana maandishi hasa ili kupata ufafanuzi wa mambo ya kale. Historia inatazama zaidi habari zilizoandikwa lakini akiolojia inatazama vitu vilivyobaki kutoka zamani. Wanaakiolojia wanaweza kutumia maandishi na habari za historia wakiamua jinsi gani waendelee na utafiti wao, kwa mfano wachimbe wapi. Lakini hutumia mitindo ya sayansi mbalimbali kuchunguza vitu vinavyopatikana kwa njia ya akiolojia.''}, the segmentation by the Syllable and BERT tokenizers are shown in Figure~\ref{fig:syllable_segments} and Figure~\ref{fig:bert_segments} respectively.

As shown in Figure~\ref{fig:syllable_segments}, the Syllable tokenizer splits words into segments based on syllables. In particular, the tokenizer splits words at each vowel. The result is that each subword contains one vowel and at least one consonant.

\subsection{GPT2 text-generation Error analysis}
We analysed the Swahili text generated by GPT-2. One guiding question was: "how natural does the sentence sound to a native Swahili speaker?". To achieve this, we evaluated the sentences generated based on four metrics, that is, language Structure, meaning, word order, and fluency. 

In Tables~\ref{tab:Example of GPT2 output on the BPE tokenizer.}, \ref{tab:Example of GPT2 output on the BERT tokenizer.}, \ref{tab:Example of GPT2 output on the Syllable tokenizer.}, we have shown one example of generated Swahili text, given the corresponding English translation and then explained why the sentence is correct. 

As shown in Table \ref{tab:Example of GPT2 output on the BPE tokenizer.}, GPT-2 was successful at keeping the correcting sentence structure and it also preserved the singularity of the sentence.

In Table \ref{tab:Example of GPT2 output on the BERT tokenizer.}, we have shown an example of an incorrect use of particles. Whereas the GPT-2 mentions ``vipaji wengi'', native Swahili speakers would say ``vipaji vingi'' instead.

The last example in Table \ref{tab:Example of GPT2 output on the Syllable tokenizer.} shows shows a correctly structured Swahili sentence in which the singularity of the sentence and the correct usage of pronouns is kept.

Lastly, we have shown human evaluation scores on 100 examples of Swahili sentences in Tables \ref{tab:Part I, Evaluation of 100 GPT2 text on language}, \ref{tab:Part II, Evaluation of 100 GPT2 text on language} and the average scores are shown in Table \ref{tab:Part II, Evaluation of 100 GPT2 text on language}. From these scores, we can see that GPT-2 performs best on ``fluency'' and least on ``language sctructure''. 
\begin{table*}[t]
    \centering
    \begin{tabular}{|p{0.2\textwidth} | p{0.4\textwidth} | p{0.4\textwidth} | }
    \hline  
      Swahili & English & Why it is correct \\
    \hline  
      Makala hii inahusu mwaka 1628 BK (Baada ya Kristo). &
      This report is about the year 1628 A.D. &
      The sentence maintains the correct sentence structure because \texttt{makala hii} refers to one entity and \texttt{inahusu} is also about one entity. Singularity is perfectly kept.\\
    \hline  
    \end{tabular}
    \caption{Example of GPT2 output on the BPE tokenizer.}
    \label{tab:Example of GPT2 output on the BPE tokenizer.}
\end{table*}
\begin{table*}[t]
    \centering
    \begin{tabular}{|p{0.33\textwidth} | p{0.33\textwidth} | p{0.33\textwidth} | }
    \hline  
      Swahili & English & Why it is incorrect \\
    \hline  
      Mchezo huu ni mojawapo kati ya majukwaa yanayotumika ku- gusa na kuupiga na kupata vipaji wengi. Mchezo unaendelea katika lugha nyingi zaidi, hasa katika uuzaji wa Kiswahili. &
      \texttt{Sentence1:} ``Mchezo huu ni mo- jawapo kati ya majukwaa yanay- otumika kugusa na kuupiga na kupata vipaji wengi.'' \texttt{Translation1:} ``This sport is one of the platforms used to touch to fight to get talents.'' \texttt{Sentence2:} ``Mchezo unaendelea katika lugha nyingi zaidi, hasa katika uuzaji wa Kiswahili.'' \texttt{Translation2:} ``This sport continues in so many languages especially in selling Swahili.'' &
      First, the \texttt{na} particle was not required. Secondly, the phrase \texttt{ vipaji wengi} should be \texttt{vipaji vingi}. This is an example of wrong usage of particles.\\
    \hline  
    \end{tabular}
    \caption{Example of GPT2 output on the BERT tokenizer.}
    \label{tab:Example of GPT2 output on the BERT tokenizer.}
\end{table*}
\begin{table*}[t]
    \centering
    \begin{tabular}{|p{0.33\textwidth} | p{0.33\textwidth} | p{0.33\textwidth} | }
    \hline  
      Swahili & English & Why it is correct \\
    \hline  
      kilimani ni kata ya wilaya ya mwanga katika mkoa wa kiliman- jaro, tanzania. wakati wa sensa iliyofanyika mwaka wa 2012, kata ilikuwa na wakazi wapatao 4, 243 walioishi humo. &
      Kilimani is a ward in Mwanga district in Kilimanjaro region, Tanzania. In 2012 there was a census that was done, Kilimani ward had approximately 4243 residents who lived there. &
      In this sentence, the object (\texttt{Kilimani} ward) is correctly described using the correct singular form and the correct pronoun all throughout the text.\\
    \hline  
    \end{tabular}
    \caption{Example of GPT2 output on the Syllable tokenizer.}
    \label{tab:Example of GPT2 output on the Syllable tokenizer.}
\end{table*}

\newpage
\newgeometry{ left = 1cm, right=1cm, top=3cm, bottom=3cm}
\begin{sidewaystable}
\centering
\scalebox{0.45}{
\begin{tabular}{|lllll|}
\toprule
Swahili Text &Language Structure &Meaning & Word order &Fluency \\
\midrule
makala hii inahusu mwaka 1878 kk ( kabla ya kristo ). &5 &5 &5 &5 \\
baadaye wafalme wawili, yakobo na yohane wa kapestrano walikusanya ujenzi huo uliokuwa umejengwa kandokando \\
ya kanisa, na kwa njia ya kanisa hilo akawa anafundisha waamini walioweza kuwaadhibu wakristo. &1 &1 &1 &4 \\
kwa jumla wastara walipata elimu ya upili katika shule mbalimbali katika shule za sekondari za wasichana za tanzania\\
na shule za msingi huko iringa, morogoro na dodoma. &2 &4 &2 &4 \\
unapatikana kaskazini mwa uganda. &5 &5 &5 &5 \\
orodha ya lugha za pakistan &5 &5 &5 &5 \\
makala hii inahusu mwaka 1898 kk ( kabla ya kristo ). &5 &5 &5 &5 \\
mnamo 26 desemba 2007, yeye pamoja na mkewe kuwa na mtoto mmoja aitwaye wawili, michael brown, walishtakiwa kwa kosa la \\
kutisha na haki ya kuripoti. siku tatu baada ya kuacha kazi, walipata nafasi ya kuanza katika kundi moja wakati wa kutoa \\
albamu yao ya kwanza mwaka wa 2009. pia walibadilishwa na kumwambia msanii mkongwe kutoka mjini new york bw. michael brown. \\
mnamo tarehe 24 aprili 2009, walitoa albamu yao ya kwanza iliyoitwa " the last king of you ". walithibitishwa na kutoa albamu \\
yake ya nne, " my super julian ". toleo la kwanza la albamu yao ya tatu, " the last king of you " ( ilitolewa septemba 2007 ), " i am.... " ". &1 &1 &1 &2 \\
kuna mbinu kadhaa kuchunguza mabadiliko katika sehemu iliyopo iliyotumika na mungu ambaye anaweza kuelewa, kuweka alama fulani \\
juu ya maneno yake, kuandika maneno yenyewe, kutumia hisia na kuelekeza. kwa mfano, mbinu ya kufuata, mbinu ya kufuata, ni dhahiri \\
kuu ya kutegemea yaliyomo katika kitabu, ambayo yalikuwa msingi wa kukosoa. mbinu ya kufuata inaweza kuwa na maana, yaani kutumia \\
maana ya maneno yenyewe. mbinu ya kufuata inaweza kuwa na lengo la kutekeleza lengo la kushiriki katika maana ya maisha, hivyo ni \\
lazima kwa watu. ni jambo la kawaida sana katika kuelewa jinsi mtu asiyetakiwa ku &1 &1 &1 &2 \\
ni korongo linalopatikana nchini burundi ( mkoa wa gitega ). &5 &5 &5 &5 \\
tangu kale wanaheshimiwa na wakatoliki na waorthodoksi kama watakatifu wafiadini. &5 &5 &5 &5 \\
mji huo ni makao makuu ya mkoa wa bahari nyekundu kusini magharibi mwa kenya. &4 &4 &4 &5 \\
mchanga mweupe &5 &5 &5 &5 \\
wakomunisti na wakomunisti. &5 &4 &4 &5 \\
maji yake yanaishia katika mto naili na hatimaye katika bahari ya kati. &5 &5 &5 &5 \\
mto kigogo ( bujumbura ) &5 &5 &5 &5 \\
mwanamke na mwanamke wa mwisho aliyeishi katika karne ya 5 au mwanzo mwa utawala wake, alijulikana kwa jina la mt. yosefu. &2 &4 &3 &4 \\
ni namba asilia inayofuata 1075 na kutangulia 1104. &5 &5 &5 &5 \\
kifo. &5 &5 &5 &5 \\
mkoa wa namibe &5 &5 &5 &5 \\
kinaga cha long 29 & & & & \\
jina linatokana na neno la kifaransa " labarage " linalomaanisha'ng'ambo ya mlima na'ng'ambo ya mlima. &4 &4 &4 &4 \\
ni mlima uliopo katika kisiwa cha ireland, upande wa jamhuri ya eire, wenye kimo cha mita 1, 298. &5 &5 &5 &5 \\
sikukuu yake huadhimishwa tarehe ya kifo chake. &5 &5 &5 &5 \\
ni kati ya mito ya mkoa wa kilimanjaro ( tanzania kaskazini mashariki ) ambayo maji yake yanaishia katika bahari hindi kupitia mto pangani. &5 &5 &5 &5 \\
kiyaberi &1 &1 &1 &1 \\
alizaliwa mwaka 1942. &5 &5 &5 &5 \\
mwaka wa 1982, bendi ya kikristo ya dar es salaam ilibadilishwa jina na kuitwa " super matiba " kwa sababu ilipokea umaarufu kutokana na wasanii wa muziki wa kizazi kipya kutoka nchi za kiarabu. &4 &5 &4 &5 \\
katika lugha ya kiebrania unatumia " anayeongea " ili kumaanisha kitu chochote ambacho kinasababisha hali mbaya ya mwisho wa dunia na hivyo matukio yale madogo yapo. &4 &5 &5 &5 \\
vigawo vyake vya namba tasa ni : 2 x 3 x 3 x 37. &5 &5 &5 &5 \\
maji yake yanaishia katika mto naili na hatimaye katika bahari ya kati. &5 &5 &5 &5 \\
mto kahuru &5 &5 &5 &5 \\
kigezo cha ukabaila. &5 &5 &5 &5 \\
kimara ni lugha ya kinilo - sahara nchini chad inayozungumzwa na wamoba. mwaka wa 2000 idadi ya wasemaji wa kimara imehesabiwa kuwa watu 150, 000. kufuatana na uainishaji wa lugha kwa ndani zaidi, kimara iko katika kundi la kimoru - madi. &5 &5 &5 &5 \\
mto kabenga ( gitega ) &5 &5 &5 &5 \\
mia saba kumi na sita ni namba inayoandikwa 299 kwa tarakimu za kawaida na dccxvi kwa zile za kirumi. &5 &4 &5 &5 \\
kilimo ni maarufu pia kwa kilimo cha mimea na majani kama cha mahindi, mpunga, ndizi, maharage, matunda, mboga mboga, viazi mbaazi n. k. ndizi hulimwa pia kwa wingi katika nchi za maziwa makuu zinazopendelea chakula, hasa katika nchi za maziwa makuu. &3 &4 &3 &4 \\
mto rwengenyi &5 &5 &5 &5 \\
kwa vyovyote mwili unaundwa na mguu ulio na magando mengi ya mbele, sehemu za uso wa sehemu yenye mafuta ya shamba au kuziba kwa fuwele. & & & & \\
kikoose ( lugha ) &5 &5 &5 &5 \\
makala hii inahusu mwaka 1201 kk ( kabla ya kristo ). &5 &5 &5 &5 \\
maisha. &5 &5 &5 &5 \\
alizaliwa kama mtoto wa nne wa baba wa benyamini akasoma sheria kwenye chuo cha kikristo mjini zurich. &5 &5 &5 &5 \\
mto kabahika &5 &5 &5 &5 \\
kiwawuli &5 &5 &5 &5 \\
\bottomrule
\end{tabular}}
\caption{Example of 100 sentences generated by GPT-2 after Syllable Tokenization. Each sentence has been evaluated on Language structure, Meaning, Word order, and Fluency on a Likert scale of 1 through 5.} \footnote{This is Part One of the Table. The second part is shown on the next page.}
\label{tab:Part I, Evaluation of 100 GPT2 text on language}
\end{sidewaystable}
\newpage
\newgeometry{ left = 1cm, right=1cm, top=3cm, bottom=2cm}
\begin{sidewaystable}
\centering
\scalebox{0.45}{
\begin{tabular}{|lllll|}
\toprule
Swahili Text &Language Structure &Meaning &Word order &Fluency \\
\midrule
baada ya vita ya wenyewe kwa wenyewe ya hispania ( 1500och hayo yalikaribishwa miaka 1000 baada ya mwisho ya mapigano ya adowa ya mwaka 1540. & &2 &3 & \\
kilimani ni kata ya wilaya ya mwanga katika mkoa wa kilimanjaro, tanzania. wakati wa sensa iliyofanyika mwaka wa 2012, kata ilikuwa na wakazi wapatao 4, 243 walioishi humo. &5 &5 &5 &5 \\
wakati ule walituma wanajeshi walioelekea nafasi kama askari. hapo mwanzo wa karne ya 19 jeshi la wajerumani lilishindwa na waingereza katika mapigano ya le 1541. &3 &3 &3 &3 \\
mto mukivuli &5 &5 &5 &5 \\
kijiji cha namtumbo &5 &5 &5 &5 \\
marejeo. &5 &5 &5 &5 \\
mto kipanga ( buvuma ) &5 &5 &5 &5 \\
kimataifa, siku moja raia wa nchi wanachama waliomfuata wakiwa wanatarajiwa kuishi duniani mpaka karne ya 13. wanajeshi walipelekwa mjini los angeles, california. kabla ya kuondolewa kwa nchi katika miaka ya 1960, kulikuwa na vita vya wenyewe kwa wenyewe ( marekani ). &4 &3 &4 &4 \\
makala hii inahusu mwaka 1064 kk ( kabla ya kristo ). &5 &5 &5 &5 \\
wakristo wanapaswa kujitoa mhanga kwa ajili ya wokovu ambao haumtaji mungu. kwa mfano wakatoliki ni wakatoliki na waprotestanti. &5 &2 &4 &4 \\
ni kati ya visiwa vya uganda kusini ( mkoa wa kati, wilaya ya kalangala ). &5 &5 &5 &5 \\
makala hii inahusu mwaka 1143 bk ( baada ya kristo ). &5 &5 &5 &5 \\
kuna zaidi ya spishi 500 za samaki huishi katika maji kame kwa vikundi pamoja na wadudu wengine. kwa kawaida aina ya kinyesi huchomwa katika samaki wa nyumba ambapo wanyama hawa ni spishi ya samaki wa baharini. wanatokea baharini kwenye maji baridi. &3 &5 &3 &4 \\
unapatikana katika kaunti ya kajiado nchini kenya ( eneo la bonde la ufa la afrika mashariki ). &5 &5 &5 &5 \\
unapatikana katika wilaya ya busia, mashariki mwa uganda. &5 &5 &5 &5 \\
mkoa wa zambezia &5 &5 &5 &5 \\
mji wake mkuu ni hohvun. &5 &5 &5 &5 \\
kwa kuwa hayajulikani mengine kuhusu historia yao, hawaorodheshwi tena na martyrologium romanum. &5 &5 &5 &5 \\
alizaliwa katika mkoa wa rukwa nchini jamhuri ya kidemokrasia ya kongo. kutokana na kazi yake kama mwanasheria, ni mwanachama wa chama cha mapinduzi. &5 &5 &5 &5 \\
unapatikana kaskazini mwa uganda. &5 &5 &5 &5 \\
mto nyamawenzi ( bujumbura ) &5 &5 &5 &5 \\
wilaya ya kyerwa ni wilaya moja ya mkoa wa kati, uganda. idadi ya wakazi wake ni takriban 2, 650, 300. &5 &5 &5 &5 \\
mto nyambu ( mwaro ) &5 &5 &5 &5 \\
maji yake yanaishia katika mto naili na hatimaye katika bahari ya kati. &5 &5 &5 &5 \\
maji yake yanaishia katika mto naili na hatimaye katika bahari ya kati. &5 &5 &5 &5 \\
unapatikana katika kaunti ya elgeyo - marakwet nchini kenya &5 &5 &5 &5 \\
wakati wa vita kuu ya pili ya dunia idadi ya askari za ujerumani iliongezeka sana ; baada ya kushindwa na wanazi na wasaliti, walipigana huko makwao na serikali ya uingereza kama ilivyosimuliwa na japani. &1 &4 &3 &4 \\
ni korongo linalopatikana nchini burundi ( mkoa wa gitega ). &5 &5 &5 &5 \\
ni mto wa mkoa wa singida ( tanzania ya kati ). &5 &5 &5 &5 \\
mnamo tarehe 10 aprili 2015, sane aliteuliwa katika kikosi cha vijana kwa klabu ya vijana wasiozidi umri wa miaka 21 ya real madrid. &5 &5 &5 &5 \\
mia sita ishirini na tatu &5 &5 &5 &5 \\
kwa kawaida ni nyeupe, lakini kadiri ya imani ya dini mbalimbali vinatenganisha maandiko matakatifu ambayo ni mapya. &2 &1 &3 &3 \\
maisha. &5 &5 &5 &5 \\
baada ya mwisho wa siasa ya marekani mwaka 1948 wamisionari wakatoliki walifaulu kuwasaidia wakazi asilia kama wakulima na wafugaji. &5 &5 &5 &5 \\
mchezaji b &5 &5 &5 &5 \\
unapatikana katika kaunti ya makueni, kusini mwa kenya. &5 &5 &5 &5 \\
wafalme wa kwanza wa uajemi ( 6 kk - 4 kk ), alikuwa mfalme wa mwisho wa babeli. &2 &4 &3 &4 \\
unapatikana kaskazini mwa burundi ( mkoa wa karuzi ). &5 &5 &5 &5 \\
ni jina la mlima ulioko katika mkoa wa morogoro nchini tanzania. &5 &5 &5 &5 \\
kimo chake ni mita 3, 127 juu ya usawa wa bahari. &5 &5 &5 &5 \\
alizaliwa katika mji wa tarrus - uliokuwa mji mkuu wa dola la ujerumani wa utamaduni wa kikoloni katika ujerumani ya kaskazini na baada ya ushindi wa wajerumani katika mapigano ya vita dhidi ya ujerumani. alichukua milki yake hadi kuporomoka kwa wajerumani na kuwa mfalme wa kwanza katika ujerumani. &1 &3 &1 &1 \\
unapatikana katika kaunti ya west pokot nchini kenya ( eneo la bonde la ufa la afrika mashariki ). &5 &5 &5 &5 \\
unapatikana kaskazini mwa uganda. &5 &5 &5 &5 \\
mia moja na ishirini ni namba inayoandikwa 303 kwa tarakimu za kawaida na cxxi kwa zile za kirumi. &5 &4 &5 &5 \\
mnamo mwaka wa 2015, mhusika mkuu wa filamu yohanne alitolea uteuzi, " ambayo pia inamiliki filamu ya uhusika wa katuni inayoongozwa na schwenzer ". &3 &1 &3 &3 \\
mto olokwaboi &5 &5 &5 &5 \\
maji yake yanaishia katika mto jubba na hatimaye katika bahari hindi. &5 &5 &5 &5 \\
historia &5 &5 &5 &5 \\
papa yohane paulo ii alimtangaza mtakatifu pamoja na wengine 119 tarehe 1 oktoba 2000. &5 &5 &5 &5 \\
alitangazwa mtakatifu na papa pius ix mwaka 1848. &5 &5 &5 &5 \\
mto nambi ( kenya ) &5 &5 &5 &5 \\
maji yake yanaelekea ziwa tanganyika, mto kongo na hatimaye bahari ya atlantiki. &5 &5 &5 &5 \\
alipofahamika, watu wa kale walishika sanamu ya kutunza urithi, tamthiliya na sanaa aliyoipata. tamthiliya na historia yake zilianza kuenea. &5 &5 &5 &5 \\
unapatikana katika wilaya ya nebbi, kaskazini mwa uganda. &5 &5 &5 &5 \\
makala hii inahusu mwaka 1165 kk ( kabla ya kristo ). &5 &5 &5 &5 \\
alimfuata papa kalisti ii akafuatwa na papa marko. &5 &5 &5 &5 \\
\toprule
\textbf{Average Scores}&4.494845361 &4.530612245 &4.520408163 &4.690721649 \\
\bottomrule
\end{tabular}}
\caption{Example of 100 sentences generated by GPT-2 after Syllable Tokenization. Each sentence has been evaluated on Language structure, Meaning, Word order, and Fluency on a Likert scale of 1 through 5.}\footnote{This is Part Two of the Table.}
\label{tab:Part II, Evaluation of 100 GPT2 text on language}
\end{sidewaystable}
\end{document}